\documentclass[pdflatex,sn-basic,iicol]{sn-jnl}

\usepackage{graphicx}%
\usepackage{multirow}%
\usepackage{amsmath,amssymb,amsfonts}%
\usepackage{amsthm}%
\usepackage{mathrsfs}%
\usepackage[title]{appendix}%
\usepackage{xcolor}%
\usepackage{textcomp}%
\usepackage{manyfoot}%
\usepackage{booktabs}%
\usepackage{algorithm}%
\usepackage{algorithmicx}%
\usepackage{algpseudocode}%
\usepackage{listings}%
\usepackage{soul} 
\usepackage{color} 
\usepackage{epstopdf}
\usepackage{xurl}

\theoremstyle{thmstyleone}%
\theoremstyle{thmstyletwo}%

\theoremstyle{thmstylethree}%

\begin{document}

\title[VAE for P-wave Detection]{Variational Autoencoders for P-wave Detection on Strong Motion Earthquake Spectrograms}


\author*[1]{\fnm{Turkan Simge} \sur{Ispak}}\email{turkan.ispak@metu.edu.tr}

\author[2]{\fnm{Salih} \sur{Tileylioglu}}\email{salih.tileylioglu@khas.edu.tr}

\author[1]{\fnm{Erdem} \sur{Akagunduz}}\email{akaerdem@metu.edu.tr}

\affil[1]{\orgdiv{Department of Modelling and Simulation, Graduate School of Informatics}, \orgname{Middle East Technical University}, \orgaddress{\city{Ankara}, \country{Türkiye}}}

\affil[2]{\orgdiv{Department of Civil Engineering}, \orgname{Kadir Has University}, \orgaddress{\city{Istanbul}, \country{Türkiye}}}


\abstract{Accurate P-wave detection is critical for earthquake early warning, yet strong-motion records pose challenges due to high noise levels, limited labeled data, and complex waveform characteristics. This study reframes P-wave arrival detection as a self-supervised anomaly detection task to evaluate how  architectural variations regulate the trade-off between reconstruction fidelity and anomaly discrimination. Through a comprehensive grid search of 492 Variational Autoencoder configurations, we show that while skip connections minimize reconstruction error (Mean Absolute Error $\approx$ 0.0012), they induce ``overgeneralization'', allowing the model to reconstruct noise and masking the detection signal. In contrast, attention mechanisms prioritize global context over local detail and yield the highest detection performance with an area-under-the-curve of 0.875. 
The attention-based Variational Autoencoder achieves an area-under-the-curve of 0.91 in the 0 to 40-kilometer near-source range, demonstrating high suitability for immediate early warning applications. These findings establish that architectural constraints favoring global context over pixel-perfect reconstruction are essential for robust, self-supervised P-wave detection.}




\maketitle

\section{Introduction}\label{sec-intro}

Earthquake Early Warning (EEW) Systems aim to detect the earliest P-wave onsets quickly and reliably in order to minimize the loss of life and infrastructure damage caused by earthquakes. Seismograph networks sense weak motion situated far from cities, while strong motion stations are placed around the cities to observe the effects in the built environment near faults \citep{molnar2017}. Strong motion sensors record higher amplitudes in noisier settings, which complicates onset detection, but they also have the potential to increase the reliability and speed of the early warning system by expanding data coverage.

Traditional methods such as Short-Term Average/Long-Term Average ($\mathit{STA/LTA}$) algorithms \citep{zhang2017} rely on predefined thresholds to detect P-wave arrivals. While effective under ideal conditions, these approaches often struggle to generalize in noisy environments or when applied to diverse seismic datasets. Moreover, as a signal processing approach, PPHASEPICKER \citep{kalkan2016} tracks the rate of change of damping energy to determine the P-wave onset. Tests have shown that it offers higher accuracy compared to the $\mathit{STA/LTA}$ method and is successful even at low signal-to-noise ratios. However, it may require specific signal processing parameters and manual verification. In contrast, machine learning methods have shown promise by leveraging data-driven models to learn intricate patterns within seismic data \citep{perol2018,mousavi2020}.

\begin{figure}[!t]
\centering
\includegraphics[width=\columnwidth]{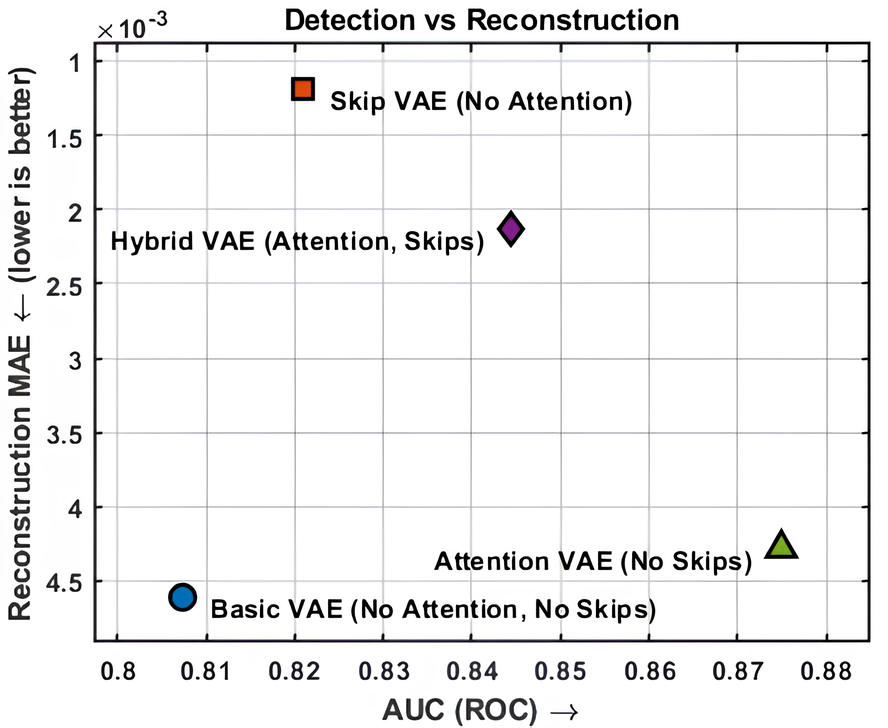}
\caption{Detection performance versus reconstruction quality for the four VAE architectures tested in our experiments. Detection is measured by the Area Under the Receiver Operating Characteristic Curve ($\mathit{AUC}$, higher is better), and reconstruction is measured by the Mean Absolute Error ($\mathit{MAE}$, lower is better). Models that reconstruct more accurately tend to detect anomalies less, effectively demonstrating a trade-off relationship.}
\label{fig:rec-det-tradeoff}
\end{figure}

In low-label-data regimes, self-supervised Variational Autoencoders (VAE) have emerged as a powerful tool for recognizing P-wave arrivals \citep{ispak2025}. Their reconstruction-driven approach captures lower-dimensional manifold representations of complex data without the need for massive labeled datasets. However, if this learned manifold becomes too expressive, the model may fail to discriminate between normal and anomalous data. 

When models overgeneralize, they can reconstruct out-of-distribution patterns (such as noise) as effectively as the target P-waves \citep{Gong2019, Angiulli2023}. This creates a fundamental trade-off between reconstruction quality and detection performance. For robust early warning, it is crucial to identify architectural elements that leverage this trade-off, forcing the model to selectively fail on anomalies to maximize the detection signal. Currently, the specific architectural inductive biases required to achieve this balance in seismic data remain largely unexplored.

In this work, we address this gap by systematically evaluating four Variational Autoencoder (VAE) architectures for P-wave detection on strong motion spectrograms. We explicitly investigate how common architectural biases, namely U-Net style skip connections and transformer self-attention bottlenecks, act as capacity controls that influence this behavior. 

We hypothesize that while skip connections favor fine detail and high-fidelity reconstruction, attention mechanisms promote global context and superior discriminative focus. By analyzing these components in the frequency domain, we aim to determine which configurations maximize the reconstruction gap between noise and P-wave arrivals, thereby directly improving detection performance. 

The architectures explored in this study are: a baseline encoder–decoder (Basic-VAE), a U-Net–style model with skip connections (Skip-VAE), a variant with attention blocks (Attention-VAE), and a hybrid combining both (Hybrid-VAE). Figure \ref{fig:rec-det-tradeoff} illustrates the resulting trade-off where the Attention-VAE achieves the best detection performance, while the Skip-VAE yield the lowest reconstruction error. The Hybrid-VAE balances these tasks, whereas the Basic-VAE serves as a baseline performance floor.

\begin{figure*}[!h]
\centering
\includegraphics[width=\textwidth]{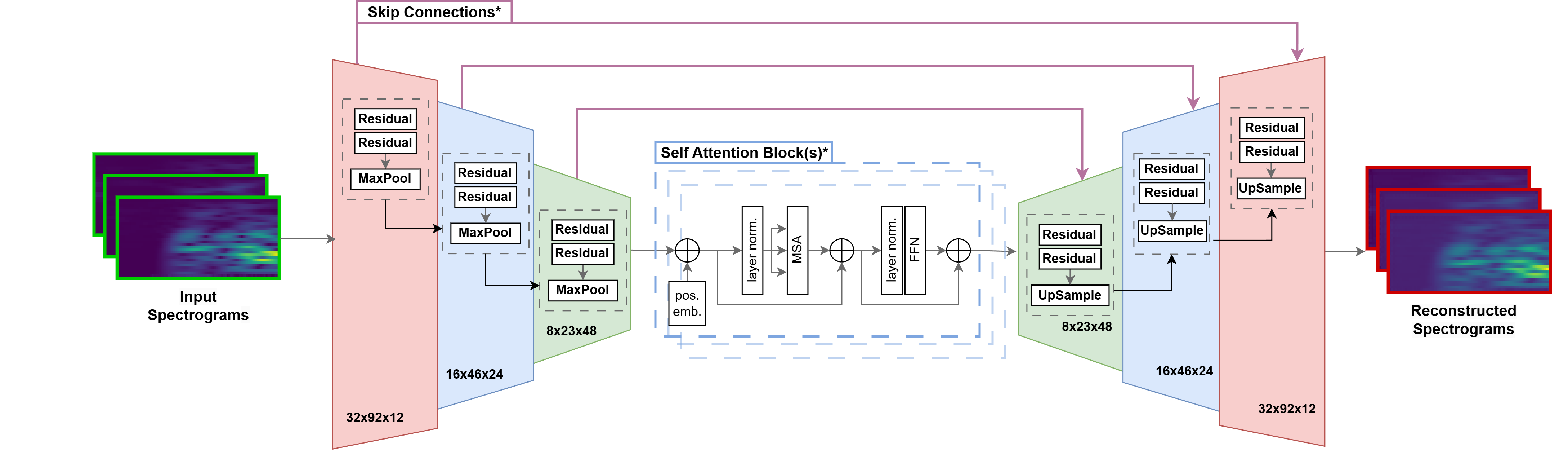}
\caption{Unified diagram of the proposed VAE architectures. Components marked with an asterisk (*) are optional, defining four configurations: Basic-VAE (baseline encoder--decoder), Skip-VAE (incorporating long-range skip connections), Attention-VAE (incorporating the self-attention bottleneck), and Hybrid-VAE (incorporating both). The encoder and decoder are symmetric, where each standard residual block consists of two convolutional layers with ReLU activation and a local skip connection.}
\label{fig:arch-hybrid-vae}
\end{figure*}

\section{Related Work}\label{sec-relatedwork}
\subsection{Deep Learning for Seismic Phase Detection}
In recent years, deep learning has emerged as the state-of-the-art for seismic phase picking, primarily through supervised learning models trained on millions of labeled seismograms. The most prominent of these is PhaseNet proposed by \citet{zhu2019}, which reformulates phase picking as an image segmentation task. PhaseNet employs a U-Net architecture introduced by \citet{ronneberger2015}, a fully convolutional encoder-decoder network, to produce probability masks for P-wave, S-wave, and noise, demonstrating remarkable performance on broadband seismic data.

Following this trend, other models have incorporated attention mechanisms, first introduced in the Transformer architecture by \citet{vaswani2017attention}, to better focus on signal onsets. \citet{ross2018} utilized a recurrent neural network (RNN) with an attention mechanism for P-wave picking and polarity determination. More recently, the Earthquake Transformer (EQTransformer) proposed by \citet{mousavi2020} adopted the full Transformer architecture, demonstrating its power for simultaneous detection and phase picking. These and other supervised methods \citep{perol2018, kim2023} have set a high benchmark for accuracy. However, their primary limitation is their reliance on large, meticulously hand-labeled datasets, which are often unavailable or difficult to curate for the specific domain of noisy, near-field strong-motion records.

\subsection{Self-Supervised Convolutional Variational Autoencoders and Anomaly Detection}
\label{vae-for-anomaly}

Unsupervised and self-supervised methods have proven success in overcoming the data-labeling bottleneck. Generative models are particularly well-suited for this task, with Autoencoders (AEs) being a common choice for learning compressed representations of data. Variational Autoencoders (VAEs), first proposed by \citet{kingma2013}, are a powerful extension. Unlike standard AEs, VAEs learn a probabilistic and regularized latent space. This probabilistic structure is a key advantage, as it encourages a continuous and generalized representation, which is highly effective for separating "normal" from "anomalous" data. For analyzing spectrograms, which are time-frequency representations, this framework is naturally extended by using convolutional encoders and decoders (Conv-VAEs) to effectively model the local dependencies and hierarchical features within the sequential data \citep{hou2017}.

This approach frames P-wave detection as a self-supervised anomaly detection problem. The model is trained as a one-class classifier, learning the underlying distribution of a single "normal" class in this case, P-wave arrivals. The model thus learns to reconstruct only the patterns of incoming P-waves. When presented with anomalous inputs (such as ambient noise or S-waves), the model, having been trained only on P-wave patterns, fails to reconstruct them accurately. Our prior work has demonstrated that this Conv-VAE framework is a viable approach, using a dual-metric score of reconstruction error and normalized cross-correlation ($\mathit{NCC}$) to robustly distinguish the "normal" P-waves (low error, high $\mathit{NCC}$) from "anomalous" noise (high error, low $\mathit{NCC}$) in strong motion data \citep{ispak2025}.

The Variational Autoencoder is often modified in various non-seismic domains to enhance performance. For instance, in industrial visual inspection \citep{collin2021}, autoencoders with U-Net-style skip connections are employed to improve reconstruction sharpness and fidelity for detecting small visual defects. In parallel, for time-series anomaly detection in fields like cybersecurity \citep{zhang2021transformer}, hybrid models integrating self-attention mechanisms are used to capture long-range, global dependencies in the data. While these architectural components are actively studied in other fields, their individual and combined impacts on seismic anomaly detection remain largely unexplored, especially within the context of P-wave detection in strong motion seismic recordings.
 
\subsection{Reconstruction-Detection Trade-off}

A critical and well-documented challenge in reconstruction-based anomaly detection, however, is that deep generative models can "generalize too well" \citep{Angiulli2023}. An autoencoder with sufficient capacity may become a powerful "pass-through" function, learning to reconstruct any input, including anomalies it has never seen. This  ``over-generalization'' capacity effectively destroys the anomaly detection capacity, as the reconstruction error for both normal and anomalous data becomes low. 

Several advanced architectures have been proposed to combat this in non-seismic-related fields. \citet{Gong2019} introduced the Memory-Augmented Autoencoder (MemAE), which forces the encoder to learn sparse, "normal" representations from a fixed memory module. While effective, this approach introduces significant model complexity and memory overhead, and the memory module itself becomes a critical hyperparameter to tune. Rather than adopting such specialized, complex architectures for seismic domains immediately, it is important to understand how foundational architectural components can be leveraged to control this delicate trade-off. 

This work specifically investigates U-Net style skip connections and Transformer-based self-attention as intrinsic capacity controls. Skip connections are known to favor high-fidelity, fine-grained reconstruction, which risks "leaking" anomalous details and worsening overgeneralization. Conversely, attention mechanisms promote a focus on global context, which we posit is better for discriminating global noise patterns from anomalous signals. While VAEs have been shown to be viable for P-wave detection \citep{ispak2025}, and this trade-off is known in general computer vision \citep{Gong2019, Angiulli2023}, the specific influence of these architectural biases on the detection-reconstruction balance for strong motion spectrograms remains unstudied. This work aims to provide a systematic investigation into this critical gap.

\begin{figure}[!t]
    \centering
    \includegraphics[width=\linewidth]{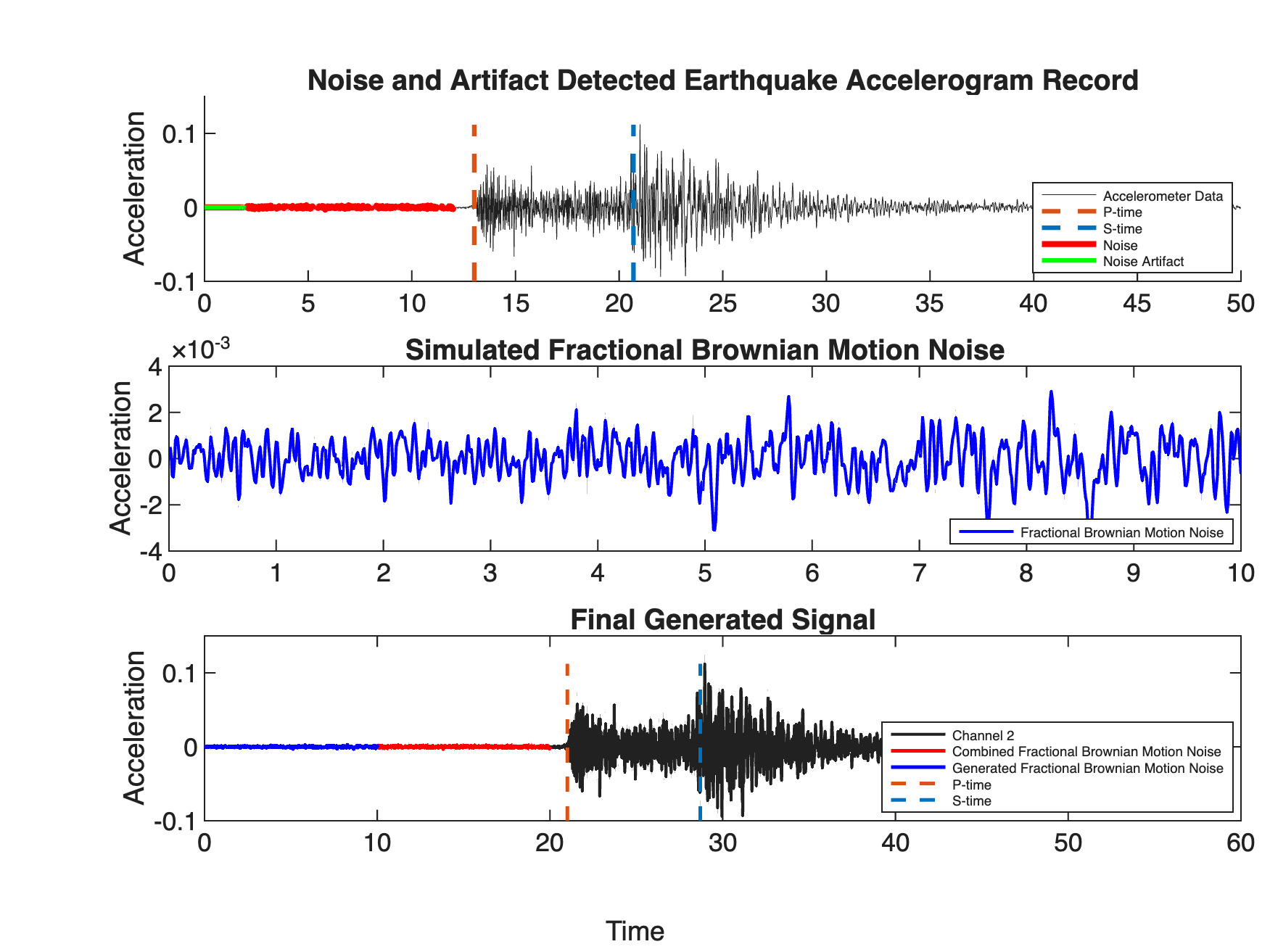}
    \caption{Original and Noise- and Artifact-Detected Earthquake Accelerogram Record. Top Panel: Original seismic record with noise and detected P and S phases. Middle Panel: Simulated FBM noise. Bottom Panel: Final augmented signal combining the original seismic data with the synthesized noise.}
    \label{fig:augmentedsignalsample}
\end{figure}

\section{Methodology}\label{sec-methodology}

\subsection{Data Acquisition and Preprocessing}
The dataset consists of three-axis strong motion accelerograms from the Turkish National Strong Motion Network (TNSMN) \citep{fdsn_tkn}, operated by the Disaster and Emergency Management Authority (AFAD) of Türkiye. The Turkish National Strong Motion Network (FDSN code TK) consists of more than 700 seismic stations deployed nationwide, each containing three-axis accelerometers. The majority are equipped with 24-bit sensors operating at a sampling rate of 100 Hz, providing high-fidelity recordings of strong-motion earthquakes. We selected 648 records containing clear seismic events and manually annotated P-wave arrival times based on established seismological criteria. To address the challenges of variable signal duration, pervasive noise, and limited labeled examples typical in strong motion data, we applied a multi-stage preprocessing pipeline.

\subsection{Artifact Removal and Noise Extraction}
Transient noise and sensor artifacts can severely degrade model training. An Adaptive Sliding Window (ASW) algorithm was devised to scan and determine the validity of the two pre-event segments. Specifically, the ASW dynamically thresholds signal amplitude to identify and remove "flatline" noise artifacts.

In operational systems, this phenomenon is identified as a buffer initialization artifact inherent to the triggered ring-buffer mechanism of digital accelerographs (e.g., GeoSIG GMS-plus, Güralp CMG-5T), also widely used in the AFAD network. As noted in the instrument specifications, the pre-event memory buffer may be incompletely filled if a trigger occurs shortly after system initialization or reset, resulting in the padding of missing samples with constant values rather than true ambient noise, as documented by \citet{geosig_manual} and \citet{guralp_manual}

\begin{figure}[!h]
    \centering
    \begin{minipage}{0.48\columnwidth}
        \centering
        \includegraphics[width=\linewidth]{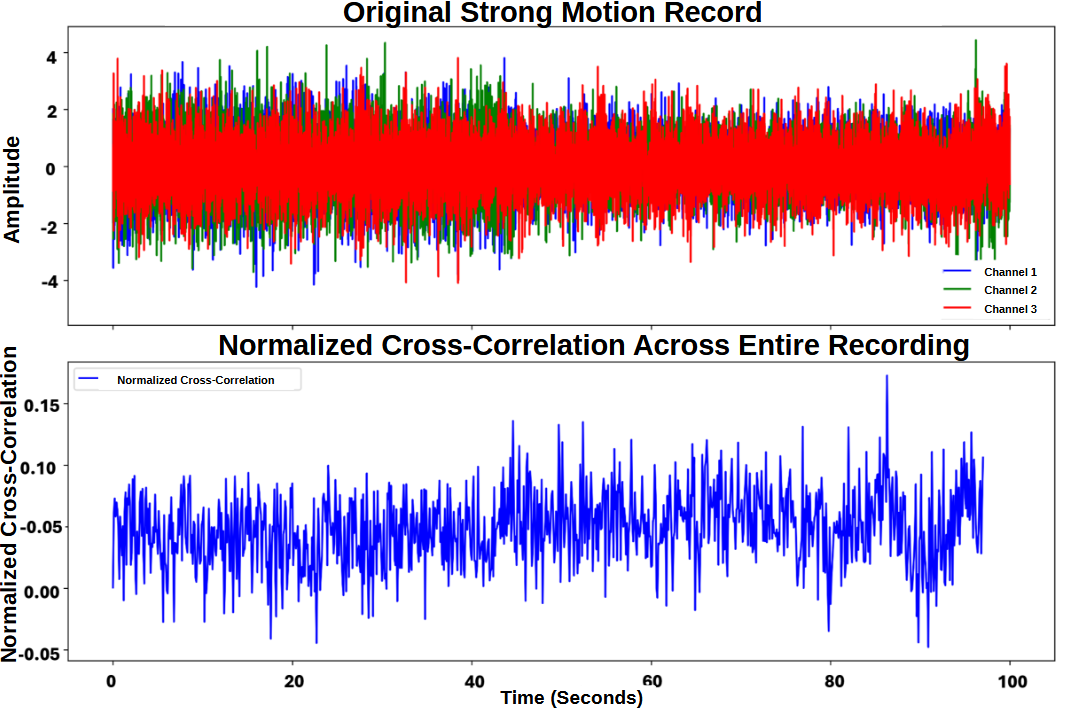}
        \centerline{\small (a) Pure Noise}
    \end{minipage}
    \hfill
    \begin{minipage}{0.48\columnwidth}
        \centering
        \includegraphics[width=\linewidth]{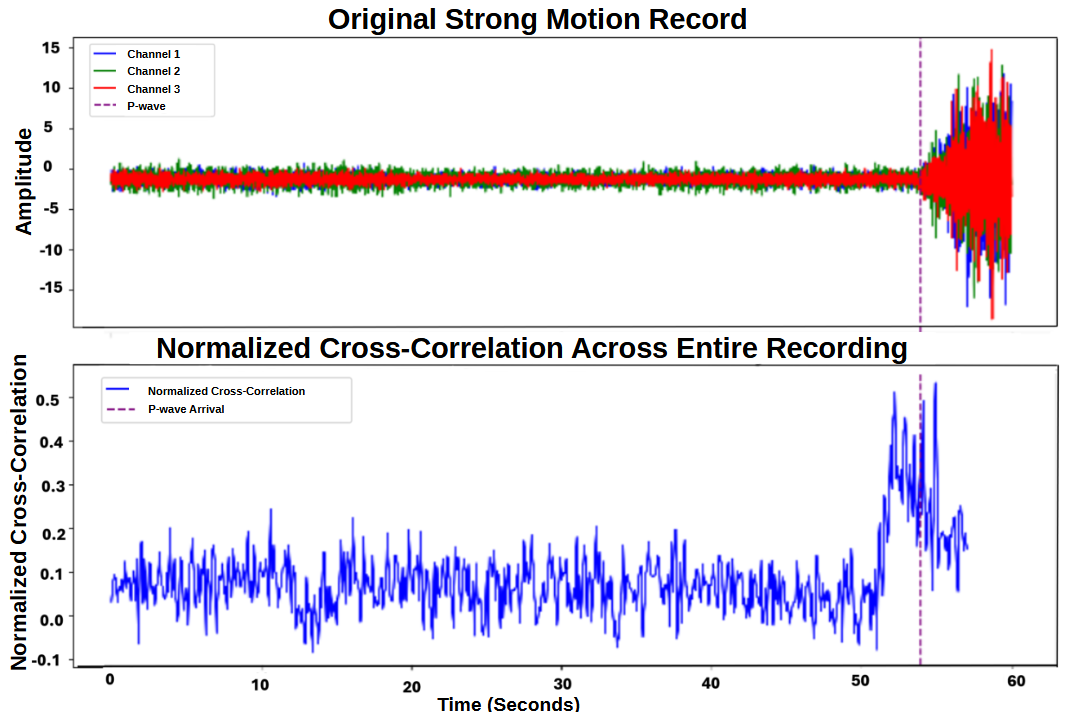}
        \centerline{\small (b) Noise and P-wave}
    \end{minipage}
    
    \vspace{1em} 
    
    \begin{minipage}{\columnwidth}
        \centering
        \includegraphics[width=\linewidth]{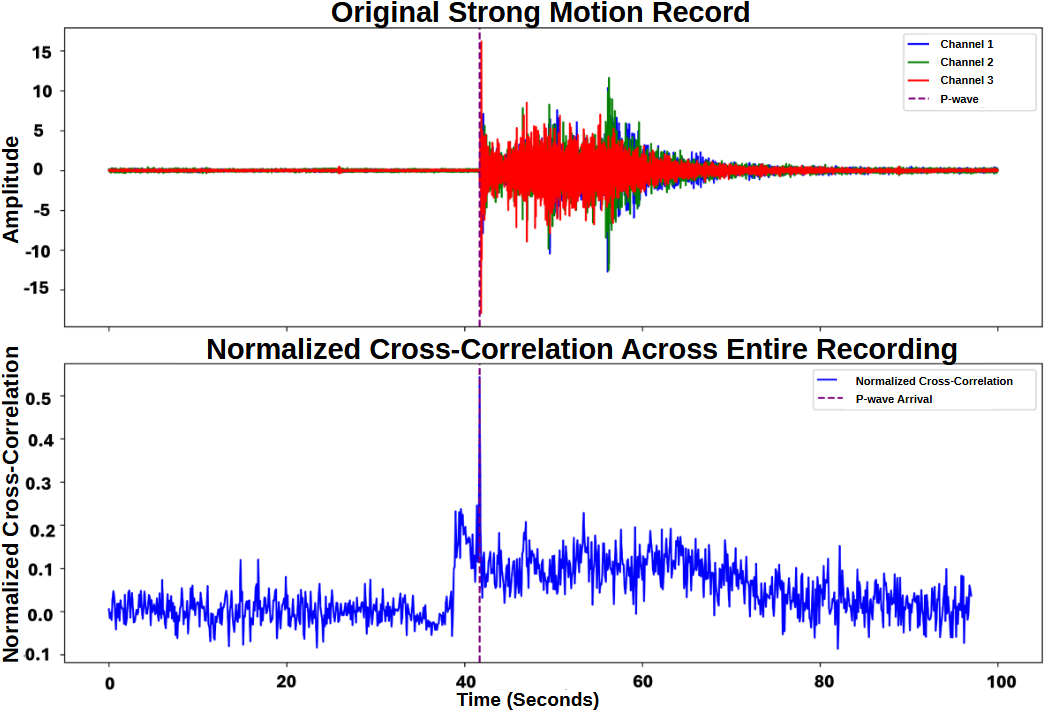}
        \centerline{\small (c) Noise, P-wave, and S-wave}
    \end{minipage}

    \caption{Normalized Cross-Correlation Analysis. (a) No significant peak is observed, minimizing false positives. (b) A distinct peak corresponding to the P-wave arrival is present. (c) Peaks are distinct for both P- and S-wave arrivals, demonstrating robustness under overlapping signals.}
    \label{fig:ncc_analysis}
\end{figure}

The literature explicitly recognizes these "noise-free pre-event samples" and typically advocates retaining them to ensure spectral compatibility during double-integration for displacement \citep{boore2005pads}, In the context of Variational Autoencoders, these non-physical "silent" segments act as out-of-distribution outliers. Retaining them would allow the model to learn "perfect silence" as a trivial feature. Although such artifacts might persist in real-time streams depending on system configuration, it is possible to remove them in the designed system. Therefore, to observe machine learning model performance independent of such mitigatable artifacts, we explicitly replaced them with characteristic noise (reflecting both site and sensor conditions).

\subsection{Characteristic Noise Augmentation}
Following the method proposed by ~\citet{zhong2018}, the objective is to generate synthetic noise whose statistical characteristics closely match those of real seismic noise. Therefore, synthetic noise is generated using Fractional Brownian Motion (FBM) to statistically match the spectral properties of the real pre-event noise extracted by the ASW. By estimating the dominant period and optimizing the Hurst exponent, we ensure the synthetic noise retains the long-range dependencies typical of pre-event seismic noise signals. This synthetic noise is prepended to real event recordings to replace any trimmed artifact segments to standardize signals to a fixed duration of 30 seconds, yielding a total of 1920 augmented samples. The complete augmentation process is illustrated in Figure \ref{fig:augmentedsignalsample}.

\subsection{Spectrogram Generation}
For model input, samples are converted from raw accelerograms into time-frequency representations. Spectrograms are preferred over raw waveforms as they explicitly localize seismic energy in both time and frequency, simplifying the separation of P-wave onsets from background noise, which often occupies distinct spectral bands. 

Each augmented record is sliced into 2.44s windows (1.00s pre-P, 1.44s post-P) with a 100ms shift. This window size balances sufficient context for discrimination with standard early-warning latency constraints. Magnitude spectra are computed via Short-Time Fourier Transform (STFT) using a 62-sample Hamming window and a 2-sample hop. The resulting [33$\times$93] spectrograms are cropped to [32$\times$92] by removing the Nyquist bin ($\sim$50Hz) and one time frame, retaining the critical $<30$Hz band where strong motion energy is concentrated. This crop ensures patch-based tokenization for architectural variants that utilize self-attention blocks. Snippets that contain a P arrival are split into train, test, and evaluation for self-supervised training. The test set includes both P and non-P snippets, while the train and evaluation contain only P arrivals.

\subsection{Generative Anomaly Detection Framework}
As introduced in subsection \ref{vae-for-anomaly}, P-wave detection is reframed as a self-supervised anomaly detection task. Given the scarcity of labeled strong-motion data that hinders standard supervised approaches, Variational Autoencoder (VAE) variants are trained exclusively on windows containing P-wave arrivals to learn a compact latent manifold of "normal" seismic onsets.

\subsubsection{Training Objective} The training objective follows a standard 
VAE formulation, minimizing a loss function composed of a mean squared reconstruction error ($\mathcal{L}_{REC}$) and a Kullback-Leibler divergence regularizer ($\mathcal{L}_{KL}$):

\begin{equation}
\mathcal{L} = \mathcal{L}_{REC}(x, \hat{x}) + \beta \cdot \mathcal{L}_{KL}(q_\phi(z|x) \| p(z))
\label{eq:vae_loss}
\end{equation}

By training only on P-waves, the model learns to reconstruct them with high fidelity. In inference, noise segments are viewed as anomalies that deviate from this learned manifold, resulting in higher reconstruction errors.

\subsubsection{Detection Mechanism} Detection is performed by scanning full-length records with a sliding window. For each window, two complementary scores are computed. The first, Mean Absolute Error ($\mathit{MAE}$), measures reconstruction fidelity and quantifies how well the expected P-wave structure is preserved. The second, Normalized Cross-Correlation ($\mathit{NCC}$), measures structural similarity independent of overall signal energy. 

The primary detection process is carried out by examining the $\mathit{NCC}$ scores, where high values and sharp increases indicate P-wave regions. High normalized cross-correlation values and sharp increases indicate the P-wave region. The magnitude of these differences in correlation scores is visualized in Figure~\ref{fig:ncc_analysis}. This visualization demonstrates that when the model indeed learns P-wave segments, it shows low correspondence (high loss) against segments that do not contain P-waves. 

Overall detection performance is measured using the area under the curve ($\mathit{AUC}$) of receiver operating characteristics ($\mathit{ROC}$) against binary (0-Noise, 1-P-wave) labels calculated using $\mathit{NCC}$ scores of each window. This metric is appropriate under class imbalance and does not require a fixed threshold. 

\subsection{Architectural Configurations and Grid Search}
All models are derived from a convolutional VAE backbone with a symmetric encoder-decoder structure, as illustrated in Figure~\ref{fig:arch-hybrid-vae}. The encoder downsamples the input to a [48, 8, 23] feature tensor, from which the latent space is derived. The four distinct architectures are obtained by selectively enabling skip connections and/or a Transformer-based bottleneck, allowing us to systematically isolate the impact of local versus global inductive biases on detection performance.

\subsubsection{Standard Variational Autoencoder (Basic-VAE)}
The baseline model is a standard convolutional VAE without skip connections or attention mechanisms. It relies solely on the variational bottleneck to compress and reconstruct the input. This configuration serves to isolate the baseline performance of the VAE framework without any specific architectural bias toward either reconstruction fidelity or discriminative focus.

\begin{figure}[h]
    \centering
    \includegraphics[width=0.9\columnwidth]{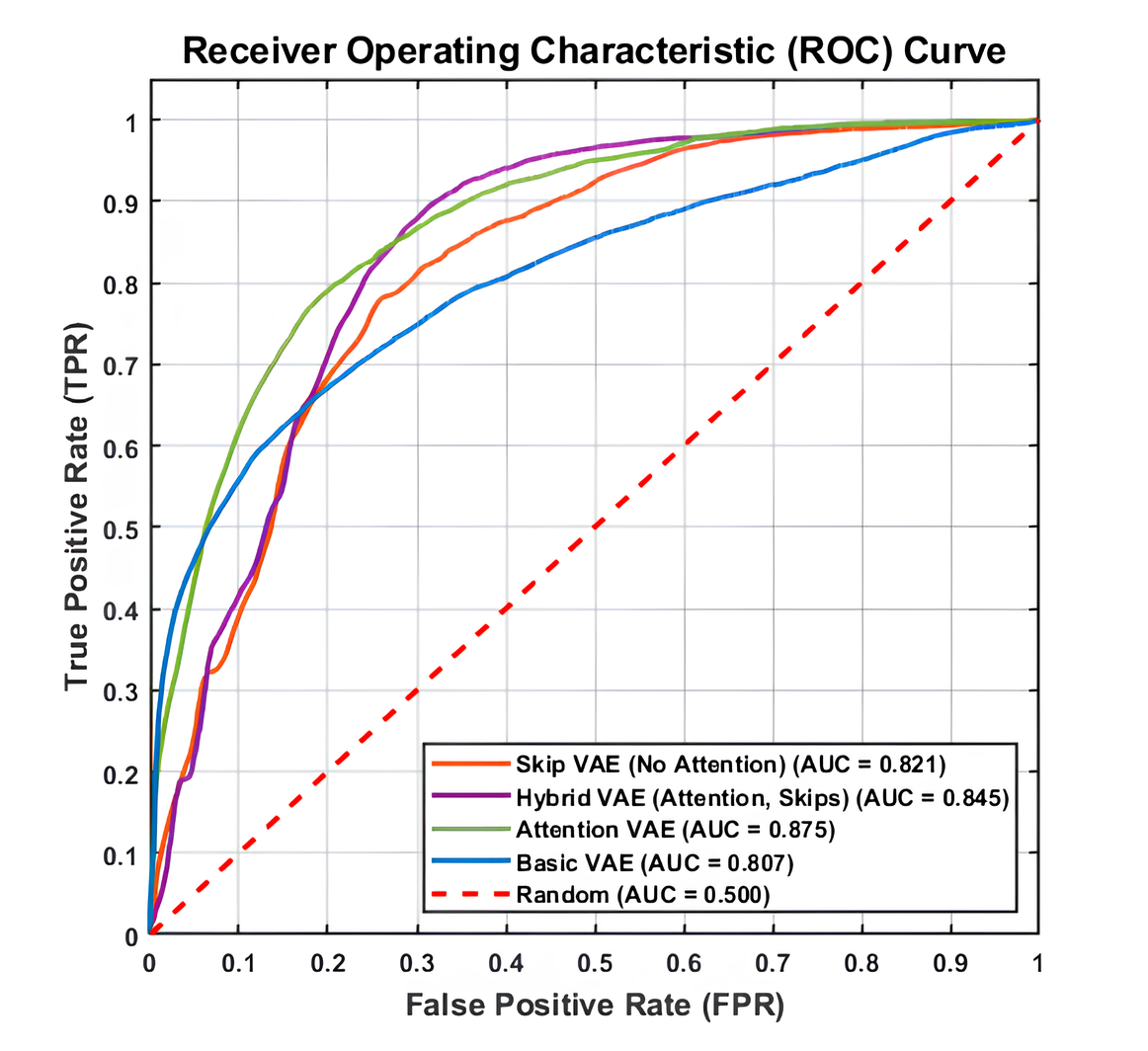}
    \caption{Comparison of Receiver Operating Characteristic ($\mathit{ROC}$) curves for the four VAE architectures. The Attention-VAE achieves the highest performance, followed by the Hybrid, Skip, and Basic models.}
    \label{fig:roc_all}
\end{figure}

\subsubsection{VAE with Skip Connections (Skip-VAE)}
This variant incorporates U-Net style skip connections. Feature maps from encoder stages are cached and concatenated with their corresponding decoder stages. The motivation is to assess the impact of these "identity shortcuts," which typically improve reconstruction quality by restoring fine-grained high-frequency details but risk reducing anomaly detection performance by allowing anomalies to bypass the latent bottleneck.

\subsubsection{VAE with Self-Attention (Attention-VAE)}
This variant replaces the standard bottleneck with a stack of Transformer-based self-attention blocks. The latent feature maps are tokenized, processed by attention layers with configurable depth and heads, and then reshaped back to [48, 8, 23] for decoding. The motivation is to leverage attention's ability to suppress irrelevant local detail and emphasize global structure, potentially enhancing onset separability at the cost of some raw reconstruction fidelity.

\subsubsection{VAE with Skip Connections and Self-Attention (Hybrid-VAE)}
The hybrid architecture combines both U-Net skip connections and the self-attention bottleneck. This design examines whether the global discriminative focus introduced by the attention mechanism can mitigate the “identity leakage” effects of local skip connections, allowing the model to sustain high detection performance while preserving reconstruction fidelity.

\subsubsection{Grid Search and Capacity Sweeps}
To map the complete performance landscape, we conduct a comprehensive grid search across these four architectures. We vary latent dimensionality (32 to 256) to control basic representational capacity. For attention-based models, we further sweep Transformer depth (1 to 48 layers) and the number of attention heads (divisors of the 48-dimensional embedding) to balance the model's ability to capture complex global dependencies against the risk of overfitting. This results in a total of 492 trained configurations, providing a vast landscape to robustly identify optimal architectural constraints.

\section{Results and Discussion}\label{sec-resultsdiscussion}
\subsection{Overview and the Detection-Reconstruction Trade-off}
From the comprehensive grid search of 492 trained configurations, the best-performing model from each architectural family was selected based on the highest Area Under the Curve ($\mathit{AUC}$), prioritizing the early warning detection objective. The performance of these four best models from each architectural variant, given in Table~\ref{tab:results}, empirically confirms the hypothesis of this work: there is a fundamental trade-off between reconstruction fidelity and anomaly detection performance in self-supervised VAEs. 

The Attention-VAE attains the best detection with a higher reconstruction error. The Hybrid-VAE follows with slightly lower $\mathit{AUC}$ and better reconstruction. The Skip-VAE achieves the best reconstruction with lower $\mathit{AUC}$. The Standard VAE is weakest overall. This ordering reveals a consistent detection and reconstruction trade-off in which architectures that emphasize context raise $\mathit{AUC}$ while those that emphasize pixel fidelity lower $\mathit{MAE}$.


\begin{table}[h]
\caption{Best-performing configuration per architecture.}\label{tab:results}
\begin{tabular}{@{}p{0.45\columnwidth}cc@{}}
\toprule
Architecture & 
\begin{tabular}[c]{@{}c@{}}Detection\\ ($\mathit{AUC}$)\end{tabular} & 
\begin{tabular}[c]{@{}c@{}}Reconstruction\\ ($\mathit{MAE}$)\end{tabular} \\
\midrule
Basic-VAE \newline (No Skips, No Attention) & 0.8074 & 0.00461 \\
\noalign{\smallskip}
Skip-VAE & 0.8210 & \textbf{0.00119} \\
\noalign{\smallskip}
Attention-VAE & \textbf{0.8749} & 0.00428 \\
\noalign{\smallskip}
Hybrid-VAE \newline (Skips and Attention) & 0.8445 & 0.00213 \\
\botrule
\end{tabular}
\end{table}

\begin{figure*}[t]
    \centering
    \includegraphics[width=\linewidth]{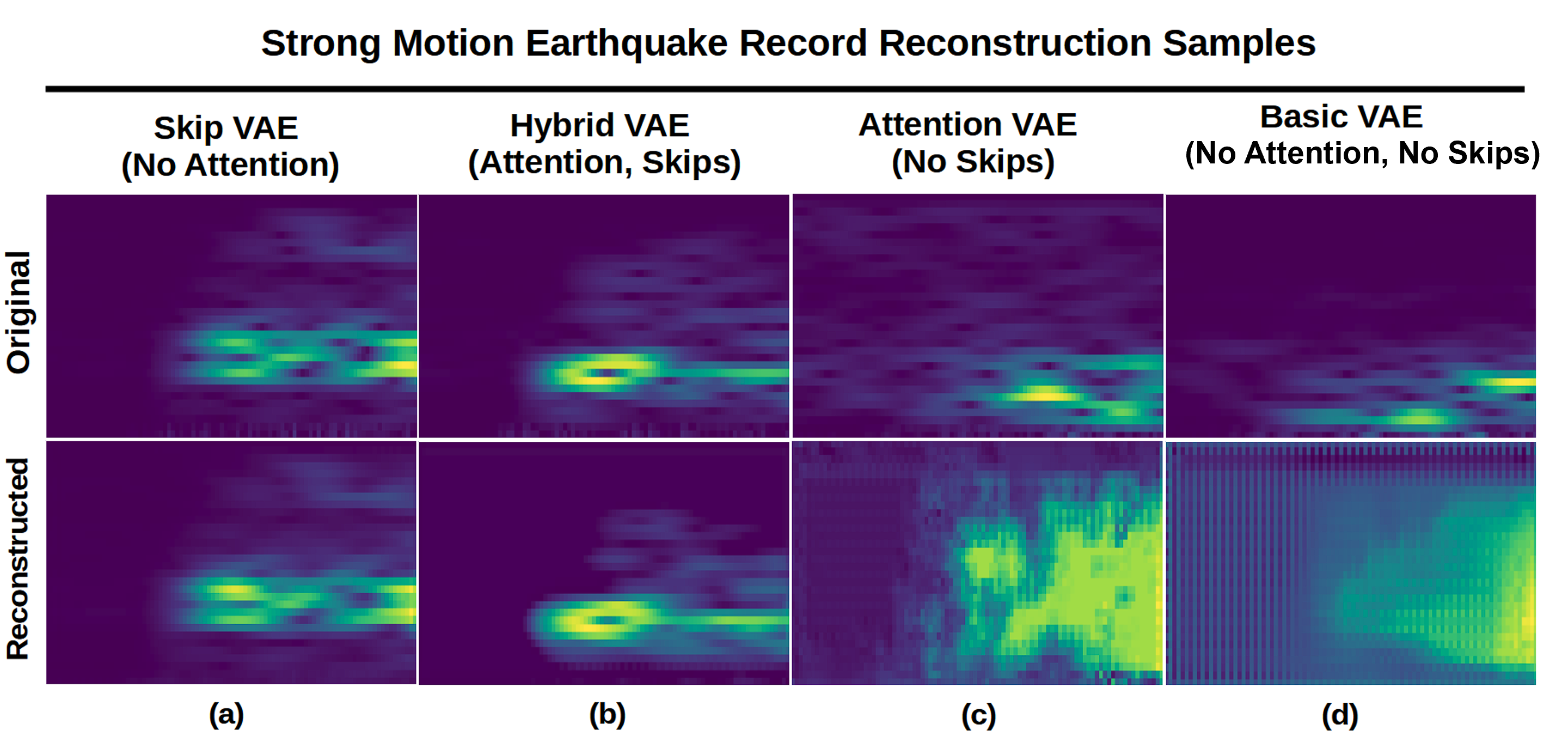}
    \caption{Reconstruction samples from each VAE architecture. (A) Skip-VAE reconstruction. (B) Hybrid-VAE reconstruction. (C) Attention-VAE reconstruction. (D) Basic-VAE reconstruction. The Skip and Hybrid models show high fidelity, while the Basic and Attention models produce blurry, generalized reconstructions.}
    \label{fig:reconsamples}
\end{figure*}

\subsection{Analysis of Detection Performance ($\mathit{AUC}$)}
Detection performance, visualized in Figure~\ref{fig:roc_all}, relies on the model's ability to generate a high-contrast Normalized Cross-Correlation ($\mathit{NCC}$) score between P-wave onsets and background noise. 

The Attention-VAE achieves the highest $\mathit{AUC}$ because its architecture is forced to learn global context. By tokenizing the latent space and utilizing self-attention, the model cannot simply copy local patterns. It must reconstruct based on the overall structure of the input. This mechanism naturally suppresses irrelevant local noise details while enhancing the separability of the "anomalous" P-wave structure, preserving a high-contrast detection signal. 

The Skip-VAE underperforms on detection due to the "overgeneralization" failure mode. While useful when high fidelity is the goal, here the direct skip pathways act as detrimental "information leaks" that bypass the variational bottleneck, creating near-identity mappings. This allows the decoder to reproduce both P-wave onsets and background noise with similarly high fidelity, compressing the $\mathit{NCC}$ gap and reducing separability.

The Hybrid-VAE keeps much of this focus while recovering some detail through skips, resulting in the second-best performance. The Skip-VAE underperforms on detection because its direct pathways let the decoder reproduce both onset and background with similar fidelity, which compresses the $\mathit{NCC}$ gap. Finally, the Basic-VAE has limited capacity for both contrast and fidelity, leading to the lowest performance with the default architecture.

\subsection{Reconstruction Quality}
Reconstruction $\mathit{MAE}$ reflects pixel-level fidelity in the spectrogram space. Sample reconstructions of the models and the trade-off between fidelity and generalization are shown in Figure~\ref{fig:reconsamples}. The Skip-VAE yields the lowest reconstruction error. Its skip connections allow for the near-perfect, high-resolution restoration of features, and its reconstructions are often almost identical to the input as seen in Figure~\ref{fig:reconsamples}.a. The Hybrid-VAE also shows strong fidelity due to its skip connections, clearly preserving the P-wave's structure and onset (Figure~\ref{fig:reconsamples}.b). However, the influence of the attention mechanism makes it slightly less of an exact copy compared to the pure Skip-VAE. 

In contrast, the Attention-VAE and Basic-VAE have higher residuals because their decoders do not receive high-resolution encoder features. This is evident in their blurry, over-smoothed reconstructions, which capture the general low-frequency energy but completely fail to reproduce the sharp, high-frequency details of the onsets (Figure~\ref{fig:reconsamples}.c-d). These models tend to generate a "generalized" or average representation, even missing the P-wave entirely in noisy segments. This visual evidence explains the detection results, models with the highest reconstruction fidelity (Skip-VAE) are poor detectors precisely because they copy both signal and noise faithfully. Models with lower fidelity (Attention-VAE) are better detectors because they effectively reconstruct "normal" noise but fail to reconstruct the "anomalous" P-wave.

\subsection{Impact of Hyperparameters on Detection Performance}
Beyond major architectural choices, the comprehensive grid search reveals critical insights into how internal capacity and hyperparameter configurations influence the stability and performance of the detection mechanism.

\subsubsection{Impact of Latent Capacity on Detection Stability}
We analyzed the sensitivity of each architecture to changes in its basic representational capacity by varying the latent dimensionality from 64 to 256. The average detection performance for each family across these capacities is presented in Table~\ref{tab:latent_impact}.

\begin{table}[h]
\centering 
\caption{Impact of Latent Dimension size on Average Detection Performance ($\mathit{AUC}$). `Fail' indicates no models in this configuration exceeded random guess performance ($>$0.5).}\label{tab:latent_impact}%
\begin{tabular}{@{}lcccc@{}}
\toprule
\begin{tabular}[c]{@{}l@{}}\textbf{Latent}\\\textbf{Dim}\end{tabular} & 
\begin{tabular}[c]{@{}c@{}}\textbf{Basic}\\\textbf{VAE}\end{tabular} & 
\begin{tabular}[c]{@{}c@{}}\textbf{Attention}\\\textbf{VAE}\end{tabular} & 
\begin{tabular}[c]{@{}c@{}}\textbf{Hybrid}\\\textbf{VAE}\end{tabular} & 
\begin{tabular}[c]{@{}c@{}}\textbf{Skip}\\\textbf{VAE}\end{tabular} \\
\midrule
64  & \textbf{0.807} & 0.630 & 0.733 & 0.514 \\
128 & 0.514 & 0.609 & \textbf{0.740} & \textbf{0.821} \\
256 & Fail & 0.551 & 0.735 & 0.791 \\
\botrule
\end{tabular}
\end{table}

The results highlight distinctive behaviors for each architecture:
\begin{itemize}
    \item {Basic-VAE (High Sensitivity):} This architecture exhibits classic signs of overgeneralization. While performing respectably at a constrained capacity (latent dimension of 64), its detection performance collapses as capacity increases. Without architectural constraints, a large latent space simply allows the model to learn an identity function, reconstructing anomalies as well as normal data.
    \item {Attention-VAE (High Variance):} Pure attention models demonstrated high sensitivity to representational capacity. Although this family yielded the highest individual $\mathit{AUC}$ (at latent dimension 128), its average performance remained inconsistent. This instability reflects the difficulty of training pure Transformers in low-data regimes, where a lack of inductive biases often leads to unstable minima without extensive tuning.
    \item {Hybrid-VAE (High Stability):} The Hybrid architecture remained stable across all tested capacities, with an average $\mathit{AUC}$ between 0.73 and 0.74. We attribute this stability to the convolutional skip connections, which provide the inductive biases needed to regularize the attention bottleneck and prevent the overgeneralization seen in the Basic-VAE.
\end{itemize}

\subsubsection{Transformer Hyperparameter Landscape}
For architectures utilizing self-attention (Attention-VAE and Hybrid-VAE), we examined the interaction between Transformer depth and the number of attention heads. A subset of this performance landscape is shown in Table~\ref{tab:transformer_hyperparams}.
\begin{figure}[h]
    \centering
    \includegraphics[width=0.9\columnwidth]{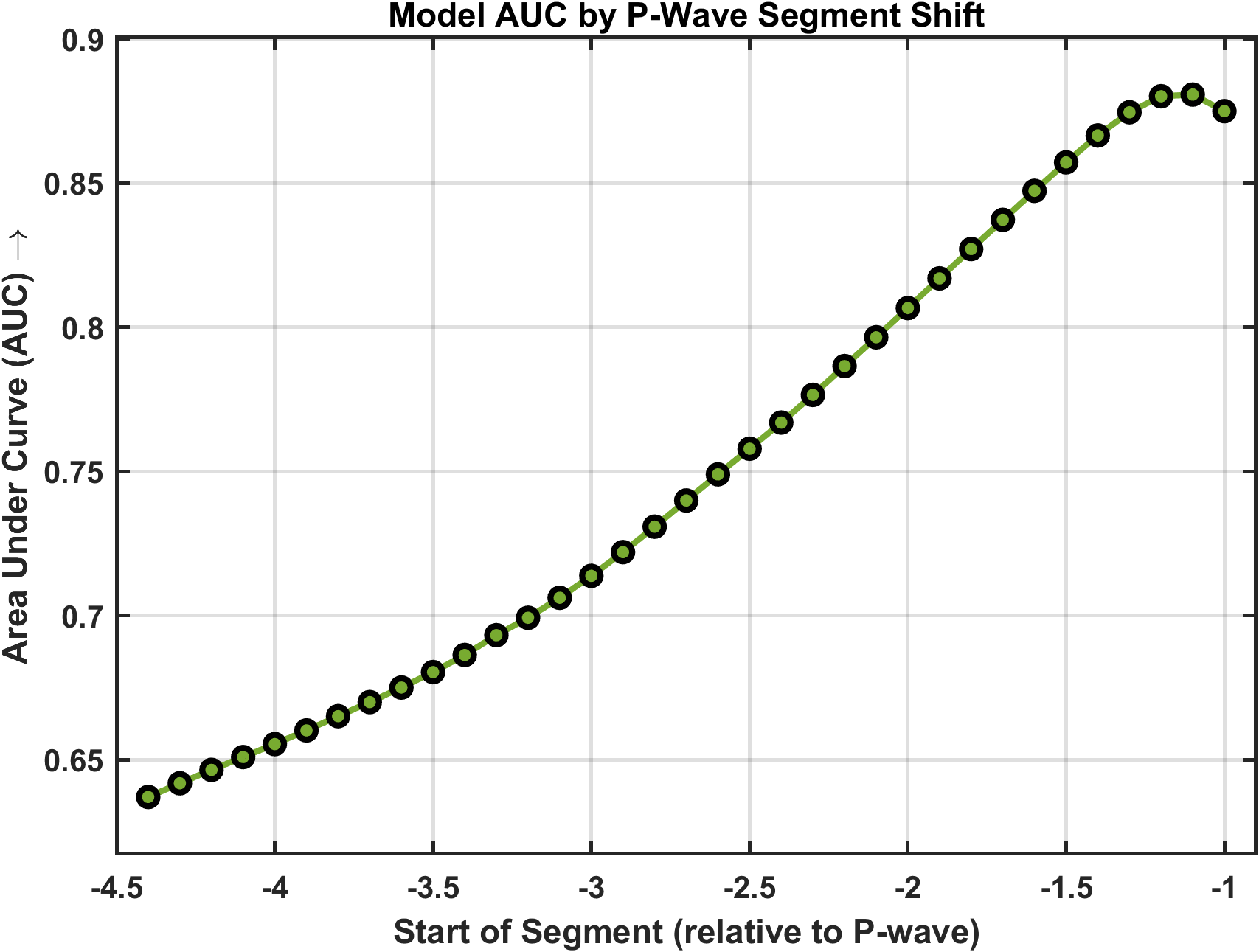}
      \caption{$\mathit{AUC}$ performance of the Attention-VAE as the 2.44s analysis window is shifted earlier relative to the P-wave onset. The detection performance declines steadily as the window moves further from the onset, indicating moderate robustness to the P-wave's arrival time.}
    \label{fig:current_shift_AUC}
\end{figure}

\begin{table}[ht]
\centering
\caption{Average $\mathit{AUC}$ for Attention-based models across varying Depth and Head configurations. An optimal range is identified at moderate-to-high complexity (16--24 layers, 16--24 heads).}\label{tab:transformer_hyperparams}
\begin{tabular}{@{}lccccc@{}}
\toprule
 & \multicolumn{5}{c}{\textbf{Attention Heads}} \\
\cmidrule(l){2-6}
\textbf{Depth (Layers)} & \textbf{4} & \textbf{8} & \textbf{16} & \textbf{24} & \textbf{48} \\
\midrule \noalign{\smallskip}
\textbf{4}  & 0.763 & 0.669 & 0.709 & 0.784 & 0.759 \\
\textbf{8}  & 0.677 & 0.758 & 0.741 & 0.752 & \textbf{0.795} \\
\textbf{16} & \textbf{0.773} & 0.709 & 0.702 & 0.730 & 0.785 \\
\textbf{24} & 0.767 & 0.746 & 0.725 & \textbf{0.825} & 0.778 \\
\botrule
\end{tabular}
\end{table}

The landscape is highly non-linear, contradicting a simple "deeper is better" hypothesis. We observe an optimal range for performance at moderate-to-high complexity, specifically with models utilizing 16 to 24 heads. This suggests that capturing complex, multi-scale global dependencies is crucial for distinguishing P-wave structure from noise. However, extremely deep models yielded diminishing returns, likely due to optimization difficulties common in very deep Transformers or overfitting to the stochastic properties of the training noise.

\subsection{Temporal Shift Sensitivity Analysis}
After determining the ideal architecture for P-wave detection as Attention-VAE, a key objective of this framework would be to quantify the model's sensitivity to the precise temporal alignment of the P-wave. To evaluate this, the optimal Attention-VAE model was subjected to a sliding window analysis in which the model's 2.44-second input window was shifted in 100\,ms increments, up to 4.5 seconds prior to the annotated P-wave onset. The resulting $\mathit{AUC}$ for each shift is plotted in Figure \ref{fig:current_shift_AUC}.

The results demonstrate a strong and clear dependency on temporal alignment. The model achieves its peak detection performance ($\mathit{AUC}$ $\approx$ 0.88) when the segment starts exactly 1.0 second before the P-wave arrival. This alignment optimally frames the P-wave onset within the 2.44-second window, just as it was designed and trained. The Attention-VAE exhibits moderate and steady degradation in performance as the window is shifted earlier. For example, shifting the window start just 1.5 seconds earlier (to -2.5s) causes the $\mathit{AUC}$ to drop to approximately 0.75. The performance degrades linearly until it reaches an $\mathit{AUC}$ of $\approx$ 0.63 at a -4.5s shift.

This steep, predictable decline validates that the model's learned features are highly specific to the P-wave's structure. The detection signal, derived from $\mathit{NCC}$, is strongest when the P-wave is correctly framed and systematically weakens as those key features are excluded. This high sensitivity indicates that the \emph{Attention-VAE} is not detecting general background noise but is highly tuned to the specific onset characteristics of the P-wave itself.

This sensitivity profile highlights a distinct behavioral divergence from our prior work \cite{ispak2025}, which utilized 1-D convolutional VAEs on raw time-series data. In that study, the 3-second 1-D models demonstrated high robustness, maintaining stable $\mathit{AUC}$ scores even when the P-wave onset was significantly shifted within the window[cite: 13, 126]. Conversely, the Attention-VAE evaluated here exhibits a linear degradation in performance as the onset moves away from the center.

This contrast suggests that while the 1-D convolutional VAEs in \cite{ispak2025} function effectively as broad `event detectors' (invariant to the signal's position), the Attention-based spectrogram model functions as a precise `structural verifier.' The attention mechanism enforces a strict reliance on the global spectro-temporal context, resulting in a model that is less forgiving of misalignment but arguably more discriminative of the specific P-wave signature. For practical EEW applications, this implies that while the Attention-VAE offers superior detection capability, it necessitates a deployment strategy with high-overlap sliding windows to ensure the onset is optimally framed.

\subsection{Detection Performance as a Function of Source--Station Distance}
\begin{figure}[t]
    \centering
    \includegraphics[width=0.85\linewidth]{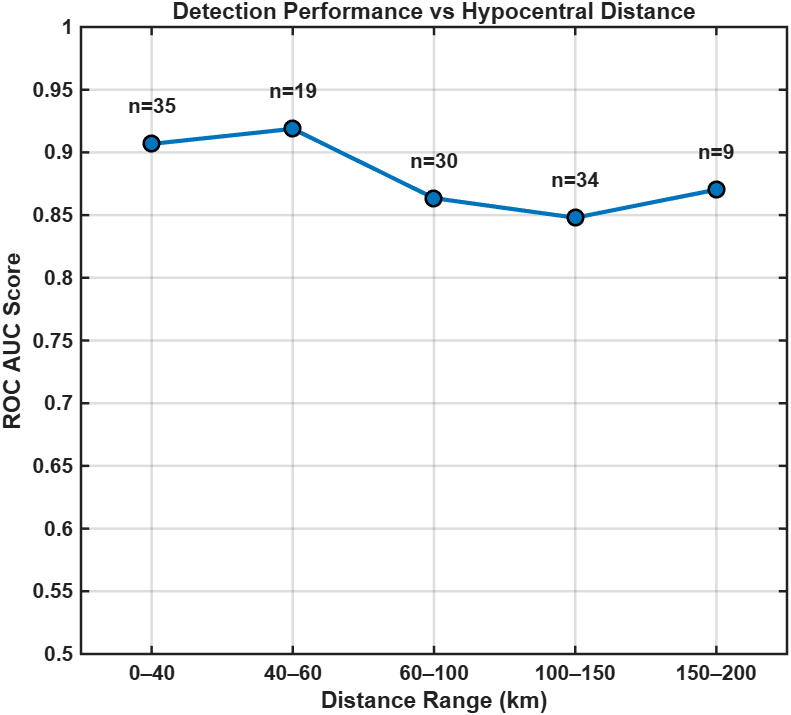}
    \caption{$\mathit{AUC}$ performance of the best Attention-VAE model across distance ranges. The annotated values ($n$) indicate the number of seismic records available in each bin.}
    \label{fig:09-AUC_by_distance}
\end{figure}

To assess how propagation effects influence self-supervised detection quality, we computed source--station distances for all events using the great-circle haversine formulation. For two points on the Earth's surface with geographic coordinates $(\phi_1,\lambda_1)$ and $(\phi_2,\lambda_2)$, the epicentral distance $\Delta$ is given by
\begin{equation}
\resizebox{0.87\columnwidth}{!}{$
\Delta = R \arccos\!\left(
\sin\phi_1 \sin\phi_2 +
\cos\phi_1 \cos\phi_2 \cos(\lambda_2 - \lambda_1)
\right)
$}
\end{equation}
where $R$ is the Earth's mean radius. The recordings were then grouped into distance bins, where $n$ denotes the sample count per bin. The $\mathit{AUC}$ performance of the best Attention-VAE model was evaluated for each group.

As shown in Fig.~\ref{fig:09-AUC_by_distance}, detection performance exhibits a systematic distance dependence. The model maintains robust performance in the near-source region, achieving an $\mathit{AUC}$ of approximately 0.91 within the 0--40~km range ($n=35$). This indicates that P-wave onsets retain the sharp, high-frequency energy most compatible with the learned manifold at shorter distances. Performance gradually decreases in the intermediate ranges (40--150~km) due to attenuation, scattering, and reduced signal-to-noise ratios. Beyond approximately 150~km, the $\mathit{AUC}$ partially recovers (0.87), potentially due to the improved large-scale coherence of low-frequency components, though we note the sample size is smaller in this range ($n=9$). Overall, this analysis confirms that the Attention-VAE maintains discriminative power across realistic operational ranges.

\section{Conclusion}\label{sec-conclusion}
In this study, we presented a systematic investigation on how architectural biases, namely U-Net style skip connections and Transformer-based self-attention, impact the performance of VAEs for P-wave detection in strong motion spectrograms. By isolating these components, we empirically validated the hypothesized trade-off between reconstruction fidelity and anomaly detection performance. Our analysis showed that standard techniques for improving signal fidelity, such as skip connections, can be actively detrimental to detection by facilitating the "overgeneralization" of background noise.

Conversely, we found that forcing the model to rely on global context through self-attention mechanisms provides the strongest discrimination for early warning, achieving the best detection performance (0.8749 $\mathit{AUC}$) despite and arguably because of its lower reconstruction fidelity. Furthermore, while pure attention models proved highly sensitive to hyperparameter capacity, the hybrid architecture demonstrated high stability, suggesting that combining convolutional inductive biases with attention can offer a more robust, if slightly less sensitive, alternative for deployment.

In conclusion, we demonstrate that prioritizing global context over local fidelity is essential for robust generative anomaly detection in seismic applications. Future work will explore several avenues to focus on making the Attention-VAE both more reliable and more practical. This includes improving training stability, expanding the dataset, and reducing the computational cost of the Transformer bottleneck through techniques such as distillation or quantization. We also plan to evaluate latency and robustness under realistic operating conditions to confirm suitability for real-time seismic monitoring.

\section*{Declarations}

\begin{itemize}
    \item \textbf{Funding:} This study was supported by the Scientific and Technological Research Council of Türkiye (TUBITAK) under Grant Number 124E066. The authors thank TUBITAK for their support.
    \item \textbf{Conflict of interest:} The authors declare that they have no competing interests.
    \item \textbf{Ethics approval:} Not applicable.
    \item \textbf{Consent to participate:} Not applicable.
    \item \textbf{Consent for publication:} All authors have read and approved the final manuscript.
    \item \textbf{Data availability:} The raw seismic records are sourced from the Turkish National Strong Motion Network (AFAD) at \url{https://tdms.afad.gov.tr/}.
    \item \textbf{Code availability:} The augmented dataset, implementation code and trained model weights are available at \url{https://github.com/turkanispak/Variational-Autoencoders-for-P-wave-Detection-on-Strong-Motion-Earthquake-Spectrograms}.
    \item \textbf{Authors' contributions:} T.S.I., E.A. and S.T. wrote the main manuscript and prepared the figures. T.S.I. performed software implementation and experimental execution. T.S.I., E.A. and S.T. developed the methodology. S.T. and E.A. acquired the funding and supervised the project. All authors discussed the results and reviewed the manuscript.
\end{itemize}

\bibliography{sn-bibliography}

\end{document}